\documentclass[11pt]{article}

\usepackage{acl}

\usepackage{times}
\usepackage{latexsym}
\usepackage[T1]{fontenc}
\usepackage[utf8]{inputenc}
\usepackage{microtype}
\usepackage{inconsolata}
\usepackage{graphicx}
\usepackage{booktabs}
\usepackage{multirow}
\usepackage{amsmath}
\usepackage{amssymb}
\usepackage{xcolor}
\usepackage{enumitem}
\usepackage{subcaption}
\usepackage{hyperref}
\usepackage{placeins}
\usepackage{float}
\usepackage{fvextra}

\newcommand{\errex}[1]{\textbf{Error Example~#1}}
\newcommand{\errexref}[1]{\hyperref[err:#1]{\errex{#1}}}

\title{RaguTeam at SemEval-2026 Task 8: Meno and Friends in a Judge-Orchestrated LLM Ensemble for Faithful Multi-Turn Response Generation}

\author{
\textbf{Ivan Bondarenko \quad Roman Derunets \quad Oleg Sedukhin} \\
\textbf{Mikhail Komarov \quad Ivan Chernov \quad Mikhail Kulakov} \\
Novosibirsk State University \\
\texttt{i.yu.bondarenko@gmail.com}
}

\begin{document}
\maketitle

\begin{abstract}
We present our winning system for Task~B (generation with reference passages) in SemEval-2026 Task~8: MTRAGEval.
Our method is a heterogeneous ensemble of seven LLMs with two prompting variants, where a GPT-4o-mini judge selects the best candidate per instance.
We ranked 1st out of 26 teams, achieving a conditioned harmonic mean of 0.7827 and outperforming the strongest baseline (gpt-oss-120b, 0.6390).
Ablations show that diversity in model families, scales, and prompting strategies is essential, with the ensemble consistently beating any single model.
We also introduce Meno-Lite-0.1, a 7B domain-adapted model with a strong cost--performance trade-off, and analyse MTRAGEval, highlighting annotation limitations and directions for improvement.
Our code is publicly available.\footnote{\href{https://github.com/RaguTeam/ragu_mtrag_semeval}
{\texttt{github.com/RaguTeam/ragu\_mtrag\_semeval}}}
\end{abstract}

\section{Introduction}
\label{sec:intro}

Retrieval-Augmented Generation (RAG) has become a standard way to ground LLMs in external evidence, but multi-turn settings remain substantially harder because systems must combine dialogue context, retrieved passages, abstention behaviour, and faithfulness. 
SemEval-2026 Task~8 MTRAGEval~\citep{rosenthal2026mtragunbenchmarkopenchallenges, Rosenthal2026MTRAGEval} addresses this gap. In Task~B, systems must answer the final user query in a multi-turn dialogue using provided reference passages.

We submit a judge-selected heterogeneous ensemble that ranks \textbf{1st} out of 26 teams. We generate candidates from seven diverse LLMs under two prompting variants and use a lightweight GPT-4o-mini judge for per-instance selection.
Our contributions are:
\begin{itemize}[nosep,leftmargin=*]
  \item \textbf{Judge-selected ensemble:} seven LLMs with two prompting strategies, selected by GPT-4o-mini (0.7827 vs.\ 0.6390 best baseline).
  \item \textbf{Diversity ablations:} gains over any single model across families, scales, and prompting strategies.
  \item \textbf{Meno-Lite-0.1:} a 7B domain-adapted model that reaches 0.681 HM$_3$ on answerable instances, demonstrating a strong cost--performance trade-off and providing occasional high-quality selections.
  \item \textbf{Benchmark analysis:} annotation and metric limitations with directions for improvement.
\end{itemize}

\section{Background}
\label{sec:related}

\paragraph{Retrieval-Augmented Generation.}
Retrieval-augmented generation (RAG) grounds language models in external evidence by combining parametric knowledge with retrieved non-parametric memory~\citep{lewis2020retrieval}. 
While early RAG systems focused mainly on single-turn, text-based question answering, recent work has extended retrieval-augmented reasoning to more heterogeneous settings. 
One line of work studies multimodal RAG, where systems retrieve and integrate evidence across text, images, tables, and other modalities, requiring cross-modal alignment and grounded generation~\citep{guo2025raganythingallinoneragframework,abootorabi-etal-2025-ask,derunets546knowledge}. 
Another line considers structured and semi-structured sources, especially tables, where retrieval must operate over schema-aware representations rather than plain passages~\citep{chen2020open,antropova-etal-2025-tabaqa,pattnayak2025hybrid}. 
Together, these directions show that RAG is increasingly moving from simple passage retrieval toward heterogeneous evidence grounding.

\paragraph{Multi-Turn RAG.}
Multi-turn RAG introduces an additional challenge: the system must interpret the current user query in the context of a dialogue history while remaining faithful to the retrieved evidence. 
This requires resolving coreferences, tracking user intent across turns, handling underspecified or unanswerable requests, and avoiding unsupported continuation from prior dialogue context. 
The mtRAG benchmark introduced a human-generated multi-turn RAG corpus with 110 conversations and 842 tasks (samples) across four domains, showing that even strong RAG systems struggle with late-turn coherence and faithful answer generation~\citep{katsis2025mtragmultiturnconversationalbenchmark}. 
MTRAG-UN further emphasises difficult cases such as unanswerable, underspecified, non-standalone, and unclear conversational turns~\citep{rosenthal2026mtragunbenchmarkopenchallenges}. 
SemEval-2026 Task~8 MTRAGEval~\citep{Rosenthal2026MTRAGEval} builds on this setting with standardised evaluation tracks; in Task~B, systems receive the dialogue history and reference passages and must generate a final-turn response grounded strictly in the provided documents.

\paragraph{Model Ensembles and LLM-Based Selection.}
LLM ensembles can exploit complementary strengths across model families, scales, and prompting strategies~\citep{jiang2023llm,wang2024mixture,lu2024merge}. 
At the same time, LLM-based evaluation provides a scalable mechanism for comparing candidate responses when reference-based metrics do not fully capture faithfulness or answer appropriateness~\citep{zheng2023judging,liu2023g}. 
Our approach combines these two ideas: we generate diverse candidate responses with a heterogeneous ensemble and use a lightweight LLM judge to select the most faithful answer for each instance.

\section{System Description}
\label{sec:system}

\begin{figure*}[t]
    \centering
    \includegraphics[width=0.75\textwidth]{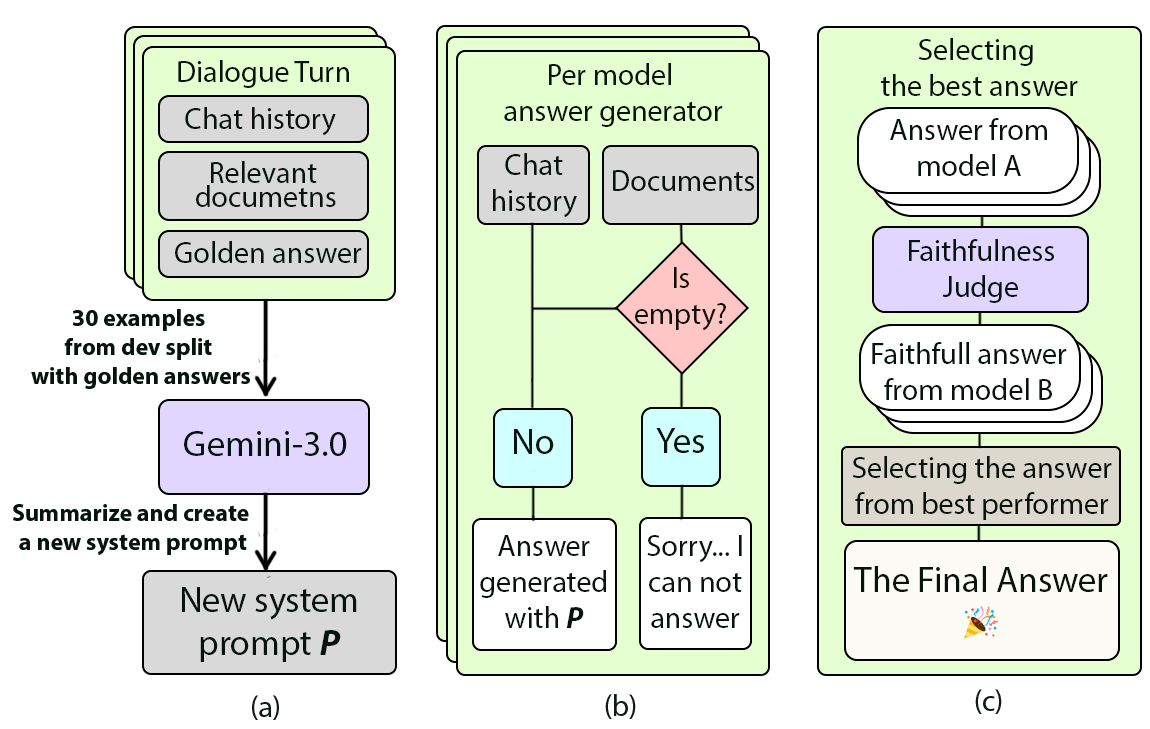}
    \caption{Overview of the proposed framework.
    \textbf{Stage~(a):} Two prompting strategies -- an iteratively refined system prompt and a category-aware few-shot variant.
    \textbf{Stage~(b):} Seven LLMs from two groups generate candidate responses.
    \textbf{Stage~(c):} GPT-4o-mini evaluates all candidates and selects the best response.}
    \label{fig:architecture}
\end{figure*}

Our system follows a three-stage pipeline illustrated in Figure~\ref{fig:architecture}: (a)~prompt construction with two distinct strategies, (b)~parallel candidate generation from seven heterogeneous LLMs, and (c)~judge-based selection of the final response.

\subsection{Prompt Engineering}
\label{sec:prompt}

\paragraph{System Prompt Design.}
We refined the system prompt iteratively using a Gemini-based procedure (detailed in Appendix~\ref{app:autodesign_of_system_prompt}): (1)~sample 30 training instances; (2)~Gemini analyses expected behaviour and synthesises a unified system prompt~$P$; (3)~evaluate $P$ on the same instances; (4)~Gemini self-critiques its responses and proposes edits; (5)~aggregate into~$P_2$. Prompt~$P$ gave small but consistent gains over a basic prompt, whereas $P_2$ degraded performance by encouraging answering despite insufficient evidence. We use~$P$ for Group~1 models. The final prompt emphasises explicit abstention for unanswerable questions, strict grounding in provided passages, concise formatting, and dialogue history use for coreference resolution (see Appendix~\ref{app:system_prompt} for the full text).

\paragraph{Few-Shot Strategy.}
For Group~2 models, we augment the prompt with category-aware few-shot examples. Training instances cluster into three structural categories: (1)~\emph{full context}: non-empty history and passages; (2)~\emph{empty context}: non-empty history but empty passages (must detect unanswerability); (3)~\emph{empty history}: non-empty passages but empty history (first-turn question). We select the exemplar from each category via medoids -- one from each of the first two categories and two from the third, yielding four exemplars total. Within each category, we embed all instances with \texttt{gte-multilingual-base}~\citep{zhang2024mgte} and choose the instance minimising total distance to others. The four exemplars improve robustness across patterns, especially for unanswerable cases (see Appendix~\ref{app:fewshot_group2} for the complete examples).

\subsection{Ensemble Composition}
\label{sec:ensemble}

Table~\ref{tab:ensemble} lists the seven ensemble models.
\textbf{Group~1} (system prompt only): Gemini-3-Pro-Preview~\citep{team2023gemini}, GLM-4.6~\citep{5team2025glm45agenticreasoningcoding}, Llama-3.3-70B-Instruct~\citep{grattafiori2024llama3herdmodels}, and Qwen3-235B-A22B-Instruct-2507~\citep{yang2025qwen3}.
\textbf{Group~2} (few-shot): Claude~4.5~Haiku (Anthropic), Qwen2.5-32B-Instruct~\citep{yang2024qwen2technicalreport}, and Meno-Lite-0.1\footnote{\href{https://huggingface.co/bond005/meno-lite-0.1}{\texttt{huggingface.co/bond005/meno-lite-0.1}}}.
The latter is a 7B model adapted from Qwen2.5-7B-Instruct via continued pretraining on Russian--English educational data and supervised fine-tuning for extraction, multi-hop QA and RAG-oriented instruction following; further details are provided in Appendix~\ref{app:meno}.
Overall, we intentionally mix providers, training pipelines and model scales to maximise diversity in failure modes.

\begin{table}[t]
\centering
\small
\begin{tabular}{@{}llcc@{}}
\toprule
\textbf{Model} & \textbf{Size} & \textbf{Type} & \textbf{Grp} \\
\midrule
Gemini-3-Pro-Preview & -- & Prop. & 1 \\
GLM-4.6 & 357B & Open & 1 \\
Llama-3.3-70B-Instruct & 70B & Open & 1 \\
Qwen3-235B-A22B-Instruct-2507 & 235B & Open & 1 \\
\midrule
Claude 4.5 Haiku & -- & Prop. & 2 \\
Qwen2.5-32B-Instruct & 32B & Open & 2 \\
Meno-Lite-0.1 & 7B & Open & 2 \\
\bottomrule
\end{tabular}
\caption{Ensemble composition. Group~1: system prompt only; Group~2: few-shot. ``Prop.'' = proprietary (API); ``Open'' = open-weight.}
\label{tab:ensemble}
\end{table}

\subsection{Judge-Based Selection}
\label{sec:judge}

For the judge, we use GPT-4o-mini, a well-known model that also served as one of the judges in the original MTRAG paper. Given seven candidate responses for each instance, GPT-4o-mini evaluates each response for \textbf{faithfulness} -- that is, whether all claims are supported by the provided passages -- and assigns a score from 0 to 1. We then select the top-ranked candidate. Ties are broken using validation-set scores.

\paragraph{Post-processing.}
For all instances where the reference context is empty, we replace the ensemble output with the fixed response ``I don't have an answer,'' ensuring reliable abstention.

\section{Experimental Setup}
\label{sec:exp_setup}

\paragraph{Dataset.}
MTRAGEval Task~B evaluation set comprises 507 instances from four collections: FiQA (77), IBMCloud (131), CLAPnq (142), and Govt (157). Answerability labels: answerable 285 (56.2\%), unanswerable 97 (19.1\%), underspecified 78 (15.4\%), partially answerable 47 (9.3\%). The training set contains 842 samples from 110 conversations, with 7.7 turns on average.

\paragraph{Metrics.}
Three metrics: \textbf{RB\_alg} (harmonic mean of BERTScore Recall~\citep{zhang2019bertscore}, BERT-K-Precision, and ROUGE-L), \textbf{RB\_llm} (LLM-judge score adapted from RAD-Bench~\citep{kuo2025rad}), and \textbf{RL\_F} (reference-less faithfulness from RAGAS~\citep{es2024ragas}). An IDK judge flags abstentions, producing conditioned variants for unanswerable questions. The official score is the harmonic mean of the three conditioned metrics (\textbf{HM$_3$}).

\paragraph{Validation.}
We use dialogue author\_id-disjoint splits: 6 authors for training (288 instances), 2 for validation (180), and 5 for testing (374). Rapid iteration uses a stratified 96-instance subset. GPT-4o-mini as a single judge yields strong agreement with organisers' multi-judge scores (Spearman $\rho{=}0.89$ unconditioned; $\rho{=}0.95$ conditioned~\citep{spearman1904proof}).

\paragraph{Implementation.}
Open-weight models run with vLLM~\citep{kwon2023efficient} on NVIDIA A100 GPUs with greedy decoding. Proprietary models are queried via APIs with default settings.

\section{Results and Analysis}
\label{sec:results}

\subsection{Main Results}
\label{sec:main_results}

\begin{table}[t]
\centering
\small
\begin{tabular}{@{}lc@{}}
\toprule
\textbf{System} & \textbf{HM$_3$ (cond.)} \\
\midrule
\textbf{Our ensemble (1st place)} & \textbf{0.7827} \\
\midrule
Baseline: gpt-oss-120b & 0.6390 \\
\bottomrule
\end{tabular}
\caption{Main results on MTRAGEval Task~B. Our system outperforms the best organiser baseline by +0.1437 absolute (+22.5\% relative).}
\label{tab:main_results}
\end{table}

Table~\ref{tab:main_results} presents our main results on the MTRAGEval Task~B final test set. Our ensemble achieves a conditioned HM$_3$ of \textbf{0.7827}, placing \textbf{1st among 26 teams} and outperforming the strongest baseline (gpt-oss-120b, 0.6390) by +14.4 absolute points.

\subsection{Individual Model Performance}
\label{sec:individual}

Table~\ref{tab:individual} reports standalone performance. The strongest individual model, GLM-4.6, achieves HM$_3{=}0.748$, while the ensemble reaches $0.783$ (organisers' judges) and $0.819$ (our single judge), confirming that judge-based selection adds value beyond any single model.
The two best models achieve their strength through different mechanisms: GLM-4.6 excels on RB\_llm, while Gemini-3-Pro-Preview leads on RL\_F -- precisely the complementarity that judge-based selection exploits.

\begin{table}[t]
\centering
\small
\resizebox{\columnwidth}{!}{
\begin{tabular}{@{}lccccc@{}}
\toprule
\textbf{Model} & \textbf{Grp} & \textbf{RB\_a$_\text{idk}$} & \textbf{RB\_l$_\text{idk}$} & \textbf{RL\_F$_\text{idk}$} & \textbf{HM$_3$} \\
\midrule
GLM-4.6 & 1 & 0.63 & 0.77 & 0.89 & 0.75 \\
Gemini-3-Pro-Preview & 1 & 0.63 & 0.75 & 0.90 & 0.74 \\
Llama-3.3-70B-Instruct & 1 & 0.63 & 0.75 & 0.86 & 0.73 \\
Qwen3-235B-A22B & 1 & 0.53 & 0.85 & 0.87 & 0.72 \\
Claude 4.5 Haiku & 2 & 0.58 & 0.76 & 0.84 & 0.71 \\
Qwen2.5-32B-Instruct & 2 & 0.48 & 0.67 & 0.72 & 0.61 \\
Meno-Lite-0.1 & 2 & 0.50 & 0.59 & 0.70 & 0.59 \\
\midrule
\textbf{Ensemble (our judge)} & -- & 0.71 & 0.82 & 0.98 & 0.82 \\
\textbf{Ensemble (MTRAGEval)} & -- & 0.64 & 0.84 & 0.93 & 0.78 \\
\bottomrule
\end{tabular}}
\caption{Individual model performance (conditioned metrics on the full evaluation set). Grp: prompting group.``Ensemble (our judge)'' reports the scores obtained during our internal testing using GPT-4o-mini as the faithfulness judge. "Ensemble (MTRAGEval)" reports the final official scores assigned by the competition organisers.}
\label{tab:individual}
\end{table}

\subsection{Ablation Study}
\label{sec:ablation}

Table~\ref{tab:ablation_all} summarises our ablation experiments across four dimensions. All values are unconditioned HM$_3$ scores; for Answerable instances, conditioned and unconditioned metrics coincide. Detailed per-metric breakdowns are in Appendix~\ref{app:model_analysis}.

\begin{table}[t]
\centering
\small
\setlength{\tabcolsep}{4pt}
\begin{tabular}{@{}llcc@{}}
\toprule
\textbf{Ablation} & \textbf{Configuration} & \textbf{ANS} & \textbf{UND} \\
\midrule
\multicolumn{4}{l}{\textit{(i) Ensemble vs.\ individual models}} \\
& Full ensemble (judge) & \textbf{.79} & .37 \\
& Best single: GLM + FS & .78 & \textbf{.38} \\
& Best single: GLM + SP & .78 & .34 \\
\midrule
\multicolumn{4}{l}{\textit{(ii) Judge vs.\ random selection}} \\
& Full ensemble (judge) & \textbf{.79} & \textbf{.37} \\
& Full ensemble (random) & .76 & .35 \\
\midrule
\multicolumn{4}{l}{\textit{(iii) Contribution of Meno-Lite-0.1}} \\
& With Meno-Lite-0.1 & .79 & \textbf{.37} \\
& Without Meno-Lite-0.1 & .79 & .36 \\
\midrule
\multicolumn{4}{l}{\textit{(iv) Few-shot (FS) vs.\ system prompt (SP)}} \\
& GLM-4.6 + FS & \textbf{.78} & \textbf{.38} \\
& GLM-4.6 + SP & .78 & .34 \\
& Llama-70B + FS & \textbf{.78} & \textbf{.38} \\
& Llama-70B + SP & .76 & .35 \\
\bottomrule
\end{tabular}
\caption{Ablation results (unconditioned HM$_3$). ANS = Answerable ($n{=}285$), UND = Underspecified ($n{=}78$). FS = few-shot prompting with basic system prompt; SP = iteratively refined system prompt without few-shot.}
\label{tab:ablation_all}
\end{table}

\paragraph{(i) Ensemble vs.\ individual models.}
On Answerable instances, the ensemble with judge selection (0.785) slightly outperforms the best standalone configuration (GLM-4.6 + few-shot, 0.782; +0.3\,p.p.). On Underspecified instances, however, the best individual model (GLM-4.6 + few-shot, 0.381) exceeds the ensemble (0.367). We attribute this to the judge occasionally selecting confident-sounding but non-clarifying responses, whereas GLM-4.6 with few-shot more consistently produces clarification requests. Across all categories combined, the ensemble outperforms every individual model due to its strength on Answerable and Unanswerable instances (where the deterministic IDK fallback achieves perfect scores).

Furthermore, the results on answerable instances alone directly refute the claim that the ensemble’s performance is largely driven by exploiting the empty-context shortcut. Even when restricted to only answerable examples, the ensemble still outperforms the best individual model (HM$_3$ = 0.785 vs. 0.782), confirming that its advantage is not solely attributable to the IDK fallback.

\paragraph{(ii) Judge vs.\ random selection.}
Judge selection consistently outperforms random selection: +2.5\,p.p.\ on Answerable (0.785 vs.\ 0.760) and +1.9\,p.p.\ on Underspecified (0.367 vs.\ 0.348). The faithfulness metric shows the largest gain (RL\_F: 0.998 vs.\ 0.922 on Answerable), confirming that the judge successfully identifies better-grounded candidates. Notably, even random selection outperforms most individual models, indicating that \textbf{diversity alone is valuable}.

\paragraph{(iii) Contribution of Meno-Lite-0.1.}
The 7B Meno-Lite-0.1 model's direct contribution to the final ensemble is limited: removing it changes Answerable and Underspecified performance by only 0.2 and 0.4 percentage points, respectively, and the judge selects it in only 2 of 424 instances. Nevertheless, these selected responses achieve strong HM$_3$ ($\approx$0.707), suggesting niche complementarity at minimal computational cost. Moreover, despite having fewer parameters than Qwen2.5-32B-Instruct, Meno-Lite-0.1 performs substantially better on Unanswerable instances in the standalone regime, achieving an IDK rate of 0.54 (see Appendix~\ref{sec:category_analysis}). Its main value is therefore cost-effective standalone performance and a useful domain-adaptation case study for this task.

\paragraph{(iv) Few-shot vs.\ iterative prompt refinement.}
Few-shot prompting with category-aware examples consistently outperforms the iteratively refined system prompt. Gains are modest on Answerable (GLM: +0.4\,p.p.; Llama: +2.3\,p.p.) but substantial on Underspecified (GLM: +3.7\,p.p.; Llama: +2.1\,p.p.), demonstrating that concrete behavioural demonstrations outperform abstract instructions, particularly for edge cases requiring clarification or abstention.

\paragraph{Summary.}
Ensemble diversity and informed selection are complementary: diversity ensures varied candidates, while the judge exploits this to select the strongest response per instance. Few-shot prompting is the most effective single intervention, especially for challenging categories.

\section{Critical Analysis of the Benchmark}
\label{sec:critical}

Through manual examination we identify several limitations (detailed in Appendix~\ref{app:benchmark}). The most consequential is \textbf{target leakage through empty context}: all 97 unanswerable questions have \emph{empty} reference passages, creating a trivial shortcut where detecting empty context and returning ``I don't know'' yields perfect conditioned scores. In realistic deployments, unanswerable questions are accompanied by \emph{irrelevant} retrieved passages, making answerability detection substantially harder.
We additionally observe annotation coverage gaps, possible metric leakage through reference-less scores computable on the submission set, and the absence of shared guidelines between generators and evaluators. We recommend including distractor passages for unanswerable instances in future iterations.

\section{Discussion and Conclusion}
\label{sec:conclusion}

\paragraph{Key Insights.}
A heterogeneous LLM ensemble with judge-based selection achieved
1st place in MTRAGEval Task~B (HM$_3{=}0.783$ vs.\ 0.639 for the
strongest baseline). Three findings generalise beyond this specific
benchmark.

\emph{First}, diversity matters more than scale: the
ensemble outperforms the best single model (GLM-4.6, 357B; HM$_3$ = 0.783
vs.\ 0.748) despite including models 5--50$\times$ smaller,
confirming that complementary failure modes outweigh raw parameter
count for grounded generation.

\emph{Second}, concrete behavioural
demonstrations outperform abstract instructions: category-aware
few-shot exemplars improve underspecified-question handling by
+3.7\,p.p.\ over an iteratively refined system prompt, suggesting
that models learn edge-case behaviour more effectively from
examples than from prescriptive directives.

\emph{Third},
faithfulness is necessary but not sufficient for pragmatic
appropriateness: the ensemble underperforms GLM-4.6 on
underspecified questions (0.367 vs.\ 0.381), motivating
multi-objective selection that balances grounding with completeness.
Beyond the ensemble, Meno-Lite-0.1 demonstrates that a
domain-adapted 7B model can match 70B-scale performance on Russian
benchmarks (MERA: 0.555, tied with Llama-3.3-70B), offering a
strong cost--performance trade-off for resource-constrained
settings.

\paragraph{System Limitations.} The following limitations apply to our system.

(i)~Running seven generators plus a judge is computationally
expensive and relies partly on proprietary APIs.

(ii)~Sequential generation and judging introduce latency unsuitable
for real-time use.

(iii)~GPT-4o-mini occasionally favours longer responses.

(iv)~Judge-based selection introduces a dependency on the judge
model; GPT-4o-mini achieves strong correlation with the organisers'
scores (Spearman $\rho{=}0.95$ conditioned), but this may not hold
for models with different failure modes, motivating evaluation of
open-weight judges and multi-judge panels.

(v)~A fully open-weight pipeline appears feasible but remains
underexplored.

\paragraph{Benchmark Limitations.}
The empty-context shortcut in MTRAGEval (all 97 unanswerable
questions have empty reference passages) incentivises a trivial
answerability detector over genuine comprehension checking, and the
reference-less faithfulness metric can be computed on the submission
set, enabling partial metric leakage. Future benchmarks should
include distractor passages for unanswerable questions and restrict
reference-less metric computation to the evaluation phase.

\paragraph{Future work.}
We plan to explore adaptive routing to invoke model subsets per
instance, including category-aware routing where a lightweight
classifier first predicts the answerability category and then routes
to specialised model subsets (e.g., Qwen3-235B for underspecified
queries, Haiku for partially answerable ones), multi-objective
judge selection combining faithfulness with completeness,
self-consistency~\citep{wang2022self} as a single-model
alternative, scaling domain adaptation to 14B--32B, and improved
abstention calibration for underspecified queries.

\section{Acknowledgments}

We thank the MTRAGEval organizers for their timely communication and the anonymous reviewers for their constructive feedback. Computational resources were provided by Novosibirsk State University.

\bibliography{custom}

\appendix

\section{Meno-Lite-0.1 Details}
\label{app:meno}

\paragraph{Design philosophy.}
Meno-Lite-0.1
is a 7B model deliberately optimised for context-grounded tasks -- RAG
question answering, information extraction, multi-hop reasoning, and
knowledge graph construction -- rather than factual recall. Its name
alludes to Plato's dialogue \emph{Meno}, in which Socrates argues
that knowledge is recollected rather than learned from scratch. We
invert this metaphor: rather than assuming knowledge is already inside
the model, we externalise it into a retrieval corpus and train the
model to ``recollect'' through a RAG pipeline. Concretely, we
hypothesise that LLM capabilities can be decomposed into \emph{world
knowledge} (facts, dates, entities) and \emph{language skills}
(comprehension, extraction, inference, generation). While world
knowledge demands ever more parameters, language skills appear to
reach a usable plateau even in 7B-class models -- provided they are
deliberately cultivated. Meno-Lite-0.1 is an empirical test of this
hypothesis: by investing training compute into language skills rather
than factual recall, we aim for a model that performs competitively on
context-grounded tasks while remaining deployable on a single consumer
GPU.

\paragraph{Model lineage.}
Meno-Lite-0.1 is derived from
RuadaptQwen2.5-7B-Lite-Beta~\citep{tihomirov2025ruadapt} through a
two-stage training pipeline. The full lineage is:
Qwen2.5-7B-Instruct $\to$ T-lite-it-1.0 $\to$ RuadaptQwen2.5-7B-Lite-Beta $\to$ Meno-Lite-0.1.
Each ancestor added a layer of Russian-language adaptation;
Meno-Lite-0.1 adds a final layer of skill-oriented training focused on
information extraction, entity normalisation, multi-hop reasoning over
long contexts, and instruction following for RAG scenarios.

\paragraph{Training data.}
\emph{Continued pretraining} (1.3B tokens): a balanced bilingual mix
of sampled FineWeb-Edu
(English)~\citep{penedo2024fineweb}, a quality-filtered RuLM subset
(Russian, selected for FineWeb-Edu similarity via
\texttt{gte-multilingual-base} embeddings), Russian educational PDFs
(FinePDFs-edu), and RuREBus~\cite{RuREBus} unlabeled text. Crucially,
English data constitutes a substantial fraction of the CPT corpus,
preventing catastrophic forgetting of English-language capabilities
acquired during the base model's original training.

\emph{Supervised fine-tuning} (50M tokens): a bilingual instruction
set designed to reinforce extraction, normalisation, summarisation, and
multi-hop QA skills. Russian-language sources include
NEREL~\citep{loukachevitch-etal-2021-nerel} NER instructions with
GPT-4o-mini-generated entity definitions and normalisations, and
LightRAG~\citep{guo-etal-2025-lightrag} query logs from Habr\footnote{\href{https://habr.com}{\texttt{habr.com}}}.
articles. English-language sources include
MultiHopRAG~\citep{tang2024multihop} multi-hop QA dialogues and
MTRAGEval training dialogues -- the same task distribution on which the
model is evaluated in this work. Additional manually created
self-cognition and alignment instructions are in Russian. The
bilingual composition ensures that RAG-oriented skills -- grounding,
abstention, cross-turn coreference -- transfer across languages rather
than being tied to a single language's surface patterns.

\paragraph{Tokenizer efficiency.}
Meno-Lite-0.1 inherits the extended vocabulary from
RuadaptQwen2.5-7B-Lite-Beta, achieving 3.77 characters per token on
Russian text -- a 47\% improvement over the original Qwen2.5 tokenizer
(2.57 chars/token) -- while maintaining English efficiency at 4.13
chars/token, on par with the best models in its class. This translates
directly into faster inference and lower serving costs, particularly
for Russian-language workloads.

\paragraph{Benchmark results.}

\emph{MERA}~\citep{fenogenova2024mera} (Russian). Meno-Lite-0.1
achieves an overall score of 0.555, matching Llama-3.3-70B-Instruct
(0.555) and substantially outperforming the base Qwen2.5-7B-Instruct
(0.482). On the MultiQ multi-hop QA subtask -- directly relevant to
RAG capability -- it scores 0.536/0.403, approaching GPT-4o
(0.572/0.431), while its world-knowledge score (CheGeKa:
0.346/0.293) reflects the deliberate trade-off in favour of
context-grounded skills over factual recall.

\emph{NEREL-bench} (knowledge graph construction; Russian).
Meno-Lite-0.1 achieves the highest harmonic mean (0.468) across entity
recognition, entity definition, relation extraction, and relation
definition, outperforming Qwen2.5-32B-Instruct (0.416) and
Gemma-3-27B-IT (0.396). This demonstrates that targeted skill
training can compensate for a 4--10$\times$ parameter disadvantage on
structured extraction tasks -- precisely the kind of capability needed
in GraphRAG and knowledge-intensive RAG pipelines.

\emph{LIBRA}~\citep{churin-etal-2025-long} (long-context
understanding; Russian). Meno-Lite-0.1 maintains near-perfect passkey
retrieval across all context lengths (1.0 at 4k--64k; 0.98 at 128k
tokens). On real-world multi-hop QA (LibrusecMHQA), it leads all
7B-class models (0.484, tied with Qwen2.5-14B-Instruct) and achieves
the highest ruQasper score at 8K context length (0.542) among all
tested models including 14B variants. These results confirm that
long-context grounding -- critical for multi-turn RAG -- does not
require large-scale parametric knowledge.

\emph{MTRAGEval} (English). In this shared task, Meno-Lite-0.1
achieves HM$_3{=}0.59$ as a standalone model (Table~\ref{tab:individual}),
substantially exceeding the organiser baseline (0.64 is the best
baseline for a much larger model). Notably, it outperforms the
4.5$\times$ larger Qwen2.5-32B-Instruct on unanswerable questions
(IDK rate of 0.54 vs.\ 0.24), suggesting that domain-specific
fine-tuning for extraction and grounded QA transfers to better
abstention calibration even in a non-Russian setting.

\paragraph{Limitations.}
As a consequence of its design, Meno-Lite-0.1 is not intended for
tasks requiring broad world knowledge without a retrieval pipeline, and
its function-calling performance is moderate (58.9\% on the BFCL
Russian benchmark, vs.\ 76.1\% for Qwen2.5-7B-Instruct). Complex
reasoning degrades at contexts beyond 32K tokens, consistent with
other models in this size class.

\section{LLM-Based Design of System Prompt for Group 1}
\label{app:autodesign_of_system_prompt}

Section~\ref{app:autodesign_of_system_prompt} describes an iterative procedure for refining the system prompt $P$ using Gemini. This appendix provides the complete prompts used at each step: (1)~analysis of agent behaviour on sampled training instances, (2)~synthesis of behavioural characteristics into a candidate system prompt, and (3)~the resulting evolution of prompt quality across iterations.

\subsection{LLM-Based Analysis of Results}
\label{app:results_analysis}

For each of the 96 sampled validation instances, we prompt Gemini to analyse the agent's behaviour along two axes: (i)~whether the response relies exclusively on the provided documents or incorporates external world knowledge, and (ii)~whether any claims are incorrect or controversial. The analysis prompt is:

\begin{Verbatim}[breaklines=true, breakanywhere=true, fontsize=\small]

Your task is to analyze the behaviour of the RAG agent with active retrieval. You are provided with a dialog between user and agent, and the documents the agent uses.

Analyze the agent behaviour: (1) does it answer based on the provided documents ONLY, of uses world knowledge beyond the provided documents?, (2) which answers are either incorrect or controversial?

The dialog:

User: What is groundwater contamination?
Assistant: Groundwater contamination refers to
harmful substances entering groundwater through
improper disposal or seepage.

User: Can it be cleaned up?
Assistant: Yes, groundwater contamination can be
cleaned up through various remediation techniques
such as pump-and-treat systems...

[... remaining dialogue turns omitted for brevity ...]

\end{Verbatim}

The output is a numbered list of behavioural characteristics per instance (e.g., ``The agent synthesises fragmented advice into declarative recommendations,'' ``It adopts a neutral persona, stripping away first-person narratives'').

\subsection{LLM-Based Creation of New System Prompt}
\label{app:new_prompt_creation}

The several per-instance behavioural lists are then fed to Gemini with a synthesis prompt that asks it to consolidate the characteristics into a concise system prompt for a 7B-scale agent. The synthesis prompt is:

\begin{Verbatim}[breaklines=true, breakanywhere=true, fontsize=\small]

I need to clone the behaviour of a specific LLM-agent in RAG scenario. Earlier I provided you with dialogs between users and the agent, where each turn the agent received the question and the documents. You analysed each case and wrote a list of the agent's behavioral characteristics.

How I will provide all the lists for all cases. Please summarize them into the instruction (system prompt) for a small 7B agent. Write around 5 list items, each of 1-2 sentences, or anything else that seems useful to simulate that agent's behavior.

The cases:

Case 1:
1. The agent synthesizes fragmented advice from the source text into declarative, actionable recommendations rather than quoting the conversational format of the documents.
2. It adopts a neutral, professional persona, stripping away first-person narratives or specific geographic examples (like "SBI" or "UK sole trader") found in the context to provide generalized advice.
3. The agent prioritizes strict adherence to the provided context, avoiding the addition of external world knowledge or hallucinations not supported by the documents.
4. It selectively extracts and combines key constraints from multiple documents (e.g., separating accounts and not mixing expenses) to form a cohesive answer.
5. The agent answers binary questions directly (starting with "Yes") and immediately supports the decision with reasoning derived specifically from the retrieved text.

Case 2:
1. The agent relies on extractive generation, lifting sentences almost verbatim from the provided documents rather than synthesizing or paraphrasing the information.
2. It concatenates excerpts from multiple documents without smoothing transitions, occasionally retaining non-narrative artifacts like document titles (e.g., "Economy of India") within the response.
3. The agent prioritizes retrieving context containing keywords over directly answering the specific question, providing related facts (like the "second largest" employer) rather than explicitly isolating the requested entity.
4. It includes tangential statistical data found in the text, such as self-employment percentages, even if this information does not strictly answer the user's specific inquiry.

[... remaining cases turns omitted for brevity ...]

\end{Verbatim}

This yields the initial system prompt $P_t$. A second round of self-critique -- where Gemini evaluates its own synthesised prompt on the same instances and proposes edits -- produces the refined prompt $P_{t+1}$.

\subsection{Step 3: Evolution of Prompt Quality}
\label{app:system_prompt_evolution}

Table~\ref{tab:prompt_evolution} traces HM$_3$ across prompt
iterations on the stratified 96-instance validation subset, using
GPT-4o-mini as the faithfulness judge. The initial bare prompt
($P_0$) provides a reasonable baseline. The LLM-synthesised prompt
 $P$ (Step~2) improves HM$_3$ by +3.6\,p.p., confirming that
extracting behavioural patterns from observed responses and encoding
them as explicit instructions yields consistent gains. However, the
self-critiqued revision $P_2$ -- which adds encouragement to
``provide the most helpful answer possible'' and to share partial
information rather than declining -- reduces HM$_3$ by 10.9\,p.p.\
relative to $P$, primarily by suppressing abstention on unanswerable
questions. We therefore adopt $P$ for all Group~1 models.

\begin{table}[t]
\centering
\small
\begin{tabular}{@{}llc@{}}
\toprule
\textbf{Version} & \textbf{Description} & \textbf{HM$_3$} \\
\midrule
 $P_0$ & Bare prompt  & 0.721 \\
 $P$   & LLM-synthesised & 0.757 \\
 $P_2$ & Self-critiqued revision of $P$            & 0.648 \\
\bottomrule
\end{tabular}
\caption{System prompt quality across design iterations on the
stratified 96-instance validation subset. HM$_3$ is estimated using
GPT-4o-mini as the faithfulness judge. $P$ is adopted for Group~1
models; $P_2$ is rejected due to degraded abstention behaviour.}
\label{tab:prompt_evolution}
\end{table}

\section{Final System Prompt for Group~1}
\label{app:system_prompt}

The following system prompt $P$ is used for all Group~1 models
(Gemini-3-Pro-Preview, GLM-4.6, Llama-3.3-70B-Instruct,
Qwen3-235B-A22B-Instruct-2507). It was produced by the
LLM-based synthesis procedure described in
Appendix~\ref{app:autodesign_of_system_prompt} and selected over
the self-critiqued revision $P_2$ based on the comparison in
Table~\ref{tab:prompt_evolution}. The five directives encode the
most consistent behavioural patterns observed across 30 sampled
training instances: strict context grounding, extractive phrasing,
neutral synthesis, concise delivery, and structural fidelity to
the source documents.

\begin{Verbatim}[breaklines=true, breakanywhere=true, fontsize=\small]
You are a helpful assistant answering questions based
strictly on provided documents. You must adhere to the
following behavioral guidelines:

1.  **Strict Context Adherence:** Rely exclusively on the
provided documents for all facts and reasoning, ignoring
external world knowledge even for common topics; if the
specific information is not in the text, do not invent it.

2.  **Extractive Phrasing:** Prioritize lifting phrases,
sentence structures, and terminology verbatim from the
source text rather than generating novel summaries; limit
paraphrasing to minor grammatical adjustments necessary
for flow.

3.  **Depersonalized Synthesis:** Combine information from
multiple document chunks into a cohesive answer, converting
first-person narratives (e.g., "I recommend") found in the
text into neutral, objective statements or second-person
advice.

4.  **Direct and Uncited Delivery:** Provide immediate,
concise answers without conversational fillers,
introductions, or meta-references (e.g., avoid "According
to the text"); for binary questions, start immediately
with "Yes" or "No" followed by the evidence.

5.  **Fidelity to Structure:** Preserve the original
formatting (such as lists) and granular technical details
from the documents, presenting them exactly as they appear
in the source material rather than simplifying or
abstracting them.
\end{Verbatim}

\section{Few-Shot Exemplars for Group~2}
\label{app:fewshot_group2}

Whereas Group~1 relies solely on a system prompt (Appendix~\ref{app:system_prompt}), Group~2 augments the simpler base prompt with four category-aware few-shot exemplars. Below we describe the exemplar selection procedure and present the chosen examples.

\paragraph{Category-aware medoid selection.}
We partition all training examples into three mutually exclusive categories based on two binary features -- whether the retrieved context is empty and whether the dialogue history is non-empty:
\begin{enumerate}[nosep]
    \item \textbf{Empty context} -- the model has no documents to ground its answer in, regardless of dialogue history;
    \item \textbf{Empty dialogue history} -- documents are available but the conversation has just started (first turn);
    \item \textbf{Non-empty context and non-empty dialogue history} -- the most common and challenging setting, requiring the model to integrate retrieved passages with prior conversational context.
\end{enumerate}
From each category we select the \emph{medoid} -- the example with the smallest total embedding distance to all other examples in the same category -- ensuring that the representative is maximally typical. Because the third category is both the largest and the most diverse, we retain two medoids from it, yielding a total of four few-shot exemplars.

\paragraph{Category~1: Empty context.}
When no documents are retrieved, the model must decline to answer rather than hallucinate. The medoid for this category demonstrates the desired fallback behaviour:

\begin{Verbatim}[breaklines=true, breakanywhere=true, fontsize=\small]
User:   There are no documents provided for the
        user's current question.
        The user's current question: Is Red Hat
        Virtualization the same as Red Hat
        OpenShift virtualization?

Assistant: The documents do not contain any
           information about Red Hat
           Virtualization to assist in deciding
           about potential differences.
\end{Verbatim}

\paragraph{Category~2: Empty dialogue history.}
When the conversation starts with documents available, the model must ground its response solely in the retrieved context. The medoid for this category illustrates a concise, document-grounded answer:

\begin{Verbatim}[breaklines=true, breakanywhere=true, fontsize=\small]
User:   The documents provided for the user's
        current question are:
        1. The document 1:
           [IBM Cloudant as a document store
            and query engine ...]
        2. The document 2:
           [IBM Cloudant databases contain JSON
            objects called documents ...]
        The user's current question: does IBM
        offer document databases?

Assistant: Yes, IBM offers IBM Cloudant which
           stores JSON objects.
\end{Verbatim}

\paragraph{Category~3: Non-empty context and non-empty dialogue history.}
This is the most frequent and challenging setting: the model must resolve co-references to prior turns (e.g., ``it'' $\to$ ``groundwater contamination'') while grounding its answer in the newly retrieved documents. We include two medoid exemplars to capture within-category diversity.

\textbf{Exemplar~3a} features a multi-turn conversation about Watson Assistant skills, where the current question (``Are dialogue skills necessary?'') requires synthesising information from the retrieved documents rather than merely echoing the conversation history:

\begin{Verbatim}[breaklines=true, breakanywhere=true, fontsize=\small]
[Conversation history]
User:   What are the different types of dialog
        nodes?
Assistant: There are various types of dialog
            nodes, including the Welcome node
            and the Anything else node [...]
User:   What are intents?
Assistant: Intents are the purposes or goals
            expressed in a customer's input [...]
User:   How is it created?
Assistant: To create a dialogue, follow these
            steps: 1. Click the Dialog tab [...]
User:   Are those the only steps?
Assistant: If there are additional steps, I am
            not aware of them as the provided
            documents do not contain any
            information about them.

[Current turn with documents]
User:   The documents provided for the user's
        current question are:
        1. The document 1:
           [Adding a skill to your assistant:
            dialog skill and search skill ...]
        2. The document 2:
           [A dialog skill has the training data
            and dialog that your assistant uses
            ...]
        3. The document 3:
           [Adding a dialog skill: the NLP for
            Watson Assistant is defined in a
            dialog skill ...]
        The user's current question: Are dialogue
        skills necessary?

Assistant: Yes. Dialog Skill contains the
           training data and dialog that your
           assistant uses to chat with customers.
           It is a container for all the
           artifacts that define a conversation
           flow. Customize your assistant by
           adding the necessary skills to
           satisfy your customers' goals.
\end{Verbatim}

\textbf{Exemplar~3b} demonstrates coreference resolution across turns (``Can \emph{it} be clean up?'' $\to$ groundwater contamination) combined with document-grounded answering:

\begin{Verbatim}[breaklines=true, breakanywhere=true, fontsize=\small]
[Conversation history]
User:   What is ground water contamination?
Assistant: Groundwater contamination refers to
            the presence of harmful substances,
            such as household cleaners, flea
            powders [...] It occurs when these
            products are disposed of improperly
            [...]

[Current turn with documents]
User:   The documents provided for the user's
        current question are:
        1. The document 1:
           [FAQs about groundwater: toxins,
            cleanup, and water fluoridation ...]
        2. The document 2:
           [FAQs continued: dos and don'ts,
            fluoridation, Garvey Reservoir ...]
        The user's current question: Can it be
        clean up?

Assistant: Yes, contaminated groundwater can be
           cleaned up, but it can be a
           time-consuming and expensive process.
           Communities whose drinking water
           source has become contaminated have
           had to spend millions of dollars to
           remove contaminants from the water
           before it can be piped to homes and
           businesses. The cleanup is effective,
           but it can significantly affect the
           cost of providing the water to
           customers. It is far better to
           prevent contamination in the first
           place.
\end{Verbatim}

% \section{System Prompt}
% \label{app:prompts}

% \todo{Insert the full text of the iteratively refined system prompt~$P$ used by Group~1 models.}

% \section{Judge Prompt}
% \label{app:judge}

% \todo{Insert the full GPT-4o-mini judge prompt, including evaluation criteria and output format instructions.}

% \section{Few-Shot Examples}
% \label{app:fewshot}

% \todo{Insert one representative few-shot example from each of the three categories: (1) full context, (2) empty context (unanswerable), (3) empty history (first turn).}

% \section{Additional Examples}
% \label{app:examples}

% \todo{Insert 2--3 qualitative examples: one good case (ensemble selects the best response), one error case (judge selects a suboptimal response), and one underspecified-question case.}

\section{Grouped Analysis}
Table~\ref{tab:group_aggregations} reports unweighted mean performance when aggregating models by prompting strategy and licensing. We intentionally used \emph{different} base models across the two prompting groups to increase architectural and behavioural diversity in the candidate pool, so that the judge can exploit per-instance complementarity rather than selecting among near-duplicate generations.   
Models using the carefully tuned \emph{system-prompt-only} setup (Group~1) outperform the \emph{system+few-shot} setup (Group~2) across all three conditioned metrics, yielding a higher average harmonic mean (0.734 vs.\ 0.634).
When grouped by availability, proprietary models (Gemini and Haiku) achieve a slightly stronger average HM$_3$ than the open-weight models (0.725 vs.\ 0.678), though this comparison is based on only two proprietary systems.
Finally, the ensemble selected by our faithfulness judge substantially exceeds the best single model (GLM-4.6), improving HM$_3$ from 0.748 to 0.819, which supports the benefit of per-instance selection over relying on any single generator. 

% \FloatBarrier

\begin{table}[!htbp]
\centering
\small
\resizebox{\columnwidth}{!}{
\begin{tabular}{@{}lccccc@{}}
\toprule
\textbf{Group (aggregation)} & \textbf{N} & \textbf{RB\_a$_\text{idk}$} & \textbf{RB\_l$_\text{idk}$} & \textbf{RL\_F$_\text{idk}$} & \textbf{HM$_3$} \\
\midrule
Prompt Grp 1: System prompt only & 4 & 0.60 & 0.78 & 0.88 & 0.73 \\
Prompt Grp 2: System + few-shot  & 3 & 0.52 & 0.68 & 0.75 & 0.63 \\
\midrule
Open-sourced models             & 5 & 0.56 & 0.73 & 0.81 & 0.68 \\
Proprietary models (Gemini+Haiku) & 2 & 0.60 & 0.75 & 0.87 & 0.73 \\
\midrule
Best single model (GLM-4.6)     & 1 & 0.63 & 0.77 & 0.89 & 0.75 \\
Ensemble (our judge)            & -- & 0.71 & 0.82 & 0.98 & \textbf{0.82} \\
\bottomrule
\end{tabular}}
\caption{Unweighted mean of conditioned metrics within each subgroup (computed over the per-model rows in Table~\ref{tab:individual}). Proprietary models include only Gemini-3-Pro-Preview and Claude 4.5 Haiku.}
\label{tab:group_aggregations}
\end{table}

\section{Analysis by Question Category}
\label{sec:category_analysis}

Table~\ref{tab:category} breaks down individual model performance by answerability category.

% \FloatBarrier

\begin{table}[!htbp]
\centering
\small
\setlength{\tabcolsep}{3.5pt}
\begin{tabular}{@{}lcccc@{}}
\toprule
\textbf{Model} & \textbf{ANS} & \textbf{UNA} & \textbf{UND} & \textbf{PAR} \\
\midrule
Gemini-3-Pro-Preview & .77 & -- & .32 & .62 \\
GLM-4.6 & .78 & -- & .33 & .65 \\
Claude 4.5 Haiku & .78 & .79 & .34 & .61 \\
Qwen3-235B-A22B-Instruct & .75 & -- & .57 & .62 \\
Qwen2.5-32B-Instruct & .78 & .24 & .37 & .63 \\
Meno-Lite-0.1 & .68 & .54 & .32 & .49 \\
\bottomrule
\end{tabular}
\caption{Conditioned $\mathrm{HM_3}$ by answerability category. ANS = answerable (285), UNA = unanswerable (97), UND = underspecified (78), PAR = partially answerable (47). The table excludes five rare selections from Llama/Qwen variants.}
\label{tab:category}
\end{table}

\paragraph{Answerable questions.}
Performance is relatively homogeneous across large models (0.750--0.779), with Claude~4.5~Haiku and Qwen2.5-32B-Instruct matching GLM-4.6.
Meno-Lite-0.1 lags at 0.681 but still exceeds the organiser baseline.

\paragraph{Unanswerable questions.}
In the final submission and in most of our evaluations, we did not run generation for questions with empty context.
Among the models evaluated on this subset, Claude~4.5~Haiku achieves 0.794 (79.4\% IDK rate), while Qwen2.5-32B-Instruct scores only 0.237 (23.7\% IDK rate), frequently generating fabricated answers when no evidence is available. Meno-Lite-0.1, despite having fewer parameters than Qwen2.5-32B-Instruct, performs substantially better on this subset, achieving an IDK rate of 0.54. This suggests that domain-specific fine-tuning can improve a model's ability to abstain when the provided context is insufficient.
This contrast motivated our deterministic empty-context fallback. Without such preprocessing, ensemble diversity would still be useful because some models abstain much more reliably than others.

\paragraph{Underspecified questions.}
This is the most challenging category for all models.
Qwen3-235B-A22B-Instruct-2507 performs best (0.566), likely because it more readily produces clarification-seeking responses.
All other models score below 0.374, suggesting that both detecting underspecification and generating appropriate clarification requests remain open challenges. The original benchmark authors similarly observe that models tend to answer eagerly by committing to a plausible but unsupported interpretation, which leads to low scores in this category even with prompts explicitly instructing the model to ask for more information when the question is underspecified (e.g., multiple plausible interpretations, overly broad scope, or missing details) \citep{rosenthal2026mtragunbenchmarkopenchallenges}.

\paragraph{Partially answerable questions.}
Models must balance providing available information while acknowledging gaps.
GLM-4.6 leads (0.649), suggesting its responses better calibrate between answering and hedging.

\section{Per-Model Contribution Analysis}
\label{app:model_analysis}

Our final ensemble selects, for each instance, the most faithful candidate using a lightweight faithfulness judge (GPT-4o-mini) on the official evaluation inputs. After submission, we analyse the organiser-reported \emph{conditioned} metrics (\texttt{RB\_alg\_idk}, \texttt{RB\_llm\_idk}, \texttt{RL\_F\_idk}) for the subset of instances for which each candidate was selected. Llama-3.3-70B-Instruct and Qwen variants were selected only in a few instances and showed no stable performance signal; we omit them for clarity. Although \texttt{Meno-Lite-0.1} was also rarely selected, it achieved strong conditioned metrics on its selected rows and is therefore reported.

\paragraph{Selection is dominated by two regimes.}
The chosen responses concentrate in (i) \textbf{ANSWERABLE} instances, where \texttt{Gemini-3-Pro-Preview} is selected most often (253/285 selected rows), and (ii) \textbf{UNANSWERABLE} instances, where a deterministic \texttt{non-model} fallback outputs an ``I do not know'' style response for empty-context cases (97 selected rows), achieving perfect conditioned scores on that subset. We excluded the \textbf{UNDERSPECIFIED} category from this analysis because the organisers did not include this subset in the leaderboard and reported zeros for the conditioned metrics.

\paragraph{PARTIAL remains the main bottleneck.}
Across answerability strata, the sharpest degradation occurs on \textbf{PARTIAL} instances: \texttt{Gemini-3-Pro-Preview} drops substantially in harmonic mean (HM) on \texttt{PARTIAL} compared to \texttt{ANSWERABLE}, indicating that incomplete evidence coverage is the primary driver of residual errors under the organiser metrics.

\paragraph{Haiku is a \emph{PARTIAL} specialist among selected rows.}
Although \texttt{Claude~4.5~Haiku} is selected infrequently overall (28 rows), it attains notably stronger HM on \textbf{PARTIAL} instances among its selected rows (HM $\approx$ 0.697 on 8 \texttt{PARTIAL} rows), outperforming \texttt{Gemini-3-Pro-Preview} on the same answerability category (HM $\approx$ 0.465 on 32 \texttt{PARTIAL} rows). This suggests that the smaller model can be more robust under underspecified evidence, potentially due to more conservative phrasing or better calibrated abstention within partially answerable queries.

\paragraph{Meno-Lite-0.1 contributes high-quality selections despite low selection rate.}
\texttt{Meno-Lite-0.1} is rarely selected (2 rows), yet achieves a strong HM on its selected instances (HM $\approx$ 0.707). While too small for firm statistical conclusions, this indicates that \texttt{Meno-Lite-0.1} can provide complementary high-quality candidates in niche cases where other models fail.

\paragraph{Implication.} These results point to \textbf{PARTIAL} handling as the most promising direction for further gains: incorporating a dedicated partial-answer policy (explicitly separating supported sub-claims from unknowns) and/or increasing the candidate pool weight of \texttt{Claude~4.5~Haiku} for low-evidence scenarios may improve \texttt{RB\_alg\_idk} without sacrificing \texttt{RL\_F\_idk}. An additional observation from the results reported in Section~\ref{sec:individual} is that \texttt{GLM-4.6} became the best single model when measured with reference-based metrics, despite not being selected most often by the single faithfulness judge. This highlights that faithfulness detection is critical but can be insufficient on its own -- especially when relying on a single judge model -- and motivates using multi-objective selection (e.g., combining faithfulness with completeness) or multiple judges to reduce selection bias. 

\begin{figure}[t]
\centering
\includegraphics[width=\columnwidth]{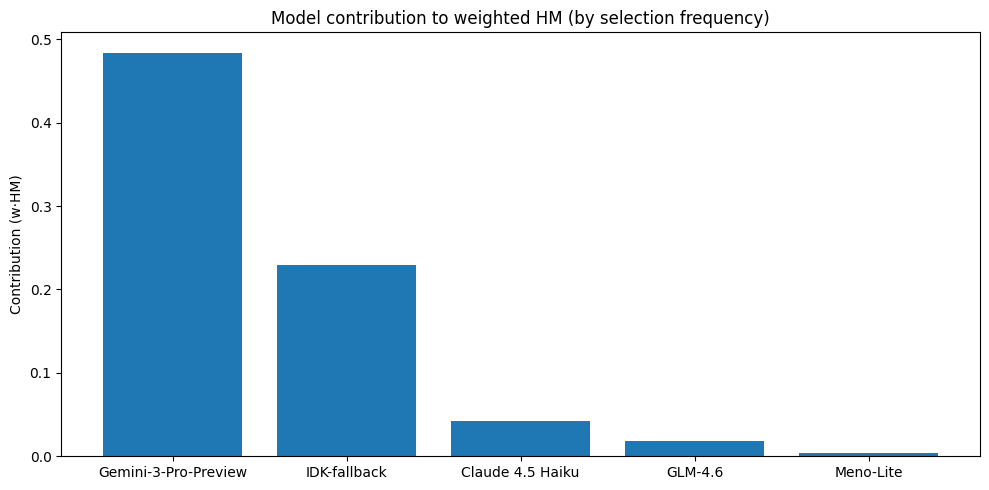}
\caption{Contribution of each model to the weighted harmonic mean of conditioned metrics. 
Each bar represents $w_i \cdot HM_i$, where $w_i$ is the fraction of instances for which the model was selected and $HM_i$ is the harmonic mean of the three conditioned metrics. Llama-3.3-70B-Instruct and Qwen family of models are omitted for clarity.}
\label{fig:model_contribution}
\end{figure}

Table~\ref{tab:per_model_selected_metrics} summarises organiser-reported conditioned metrics for instances where each model was selected by our faithfulness judge. To better understand how individual models contribute to the final system performance, we compute a contribution-weighted harmonic mean, where each model's harmonic mean over the three conditioned metrics is scaled by its selection frequency. Formally, for model $i$ with $N_i$ selected instances and harmonic mean $HM_i$, we define its contribution as $w_i \cdot HM_i$, where $w_i = N_i / \sum_j N_j$. Figure~\ref{fig:model_contribution} visualises these contributions. 

% \FloatBarrier

\begin{table*}[!htbp]
\centering
\small
\setlength{\tabcolsep}{4pt}
\resizebox{\textwidth}{!}{
\begin{tabular}{lrrrrrrrr}
\hline
\textbf{Model} & \textbf{N} & \textbf{Ans.} &
\textbf{RB\_alg$_\text{idk}$} & \textbf{RB\_llm$_\text{idk}$} & \textbf{RL\_F$_\text{idk}$} &
\textbf{HM$_{(3)}$} & $\boldsymbol{w}$ & $\boldsymbol{w}{\cdot}\boldsymbol{HM}$ \\
\hline
Gemini-3-Pro-Preview & 285 & all & 0.55 & 0.80 & 0.93 & 0.72 & 0.67 & 0.48 \\
\quad & 253 & ANS & 0.57 & 0.84 & 0.96 &  &  &  \\
\quad & 32 & PAR & 0.34 & 0.51 & 0.64 &  &  &  \\
\hline
Claude~4.5~Haiku & 28 & all & 0.46 & 0.75 & 0.87 & 0.65 & 0.07 & 0.04 \\
\quad & 20 & ANS & 0.45 & 0.73 & 0.84 &  &  &  \\
\quad & 8 & PAR & 0.49 & 0.83 & 0.94 &  &  &  \\
\hline
GLM-4.6 & 12 & all & 0.48 & 0.75 & 0.95 & 0.67 & 0.03 & 0.02 \\
\quad & 7 & ANS & 0.50 & 0.86 & 0.91 &  &  &  \\
\quad & 5 & PAR & 0.44 & 0.59 & 1.00 &  &  &  \\
\hline
Meno-Lite-0.1 & 2 & ANS & 0.52 & 0.75 & 1.00 & 0.72 & 0.01 & 0.00 \\
\hline
IDK-fallback & 97 & UNA & 1.00 & 1.00 & 1.00 & 1.00 & 0.23 & 0.23 \\
\hline
\multicolumn{8}{r}{\textbf{Sum of contributions} $\sum_i w_i HM_i$} & \textbf{0.78} \\
\hline
\end{tabular}}
\caption{Organiser conditioned metrics for instances where each candidate was selected by our faithfulness judge (GPT-4o-mini). $N$ is the number of selected instances; ANS/PAR/UNA denote answerability; \textit{IDK-fallback} is a deterministic ``I do not know'' output for empty context. We also report the selection weight $w_i=N_i/N_{\text{tot}}$ ($N_{\text{tot}}{=}424$) and the contribution $w_i\!\cdot\!HM_i$, where $HM_i$ is the harmonic mean of the three conditioned metrics.}
\label{tab:per_model_selected_metrics}
\end{table*}
% \FloatBarrier

\section{Additional Preprocessing Steps}
\label{sec:coref}

In this section, we describe two optional preprocessing components that were explored during development but not used in the final submission.

\paragraph{Query Rewriting.}
As an optional preprocessing step, we explored a dedicated coreference-resolution agent that rewrites the last user utterance into a self-contained query by (i)~replacing relative temporal expressions (e.g., ``last year'', ``recently'') with concrete calendar dates, and (ii)~resolving pronominal and nominal references (e.g., ``he'', ``the company'') using the dialogue context.  
The exact prompt used for this agent is shown below:

\begin{Verbatim}[breaklines=true, breakanywhere=true]
You are a coreference resolution agent.
Today's date is {date}.

Your task:
Replace relative temporal expressions
(e.g., 'last year', 'this year',
'recently', 'now') with explicit
calendar dates or years relative to
today's date. Deduce mentions like
'he', 'she', 'they', 'the company', etc.
from the context of the conversation.

Rewrite the last user message with all
coreferences and temporal expressions
resolved. Do not answer the question,
just rewrite it.
\end{Verbatim}

We included four few-shot demonstration rounds structured as prior user–assistant exchanges, following the in-context learning paradigm~\citep{brown2020language}. One illustrative example is shown below:

\begin{Verbatim}[breaklines=true, breakanywhere=true]
User: What is groundwater contamination?
Assistant: Groundwater contamination 
refers to harmful substances entering
groundwater through improper disposal
or seepage.

User: Can it be clean up?
Assistant (rewritten query):
Can groundwater contamination
be cleaned up?
\end{Verbatim}

However, this step did not improve generation performance for Qwen3-8B (conditioned HM: 0.615 vs.\ 0.619 without rewriting). The combination of dialogue history and reference passages in Task~B already provides sufficient disambiguating context at generation time. We therefore did not apply query rewriting in the final submission, although it remains well-motivated for retrieval-stage improvements~\citep{sun-etal-2023-improving, katsis2025mtragmultiturnconversationalbenchmark}.

\paragraph{Relevance Filtering.}
We also implemented a relevance agent that iterates over retrieved documents and predicts whether each document is relevant to the user's question. The prompt used for this binary classification is shown below:

\begin{Verbatim}[breaklines=true, breakanywhere=true]
Determine whether the document is
relevant to the user question.

Answer ONLY "yes" or "no".
\end{Verbatim}

In Task~B, however, all provided reference chunks are already considered relevant, rendering this step largely redundant.

\section{Error Analysis}
\label{sec:errors}

We manually analysed errors in our final submission, identifying five dominant failure patterns:

\paragraph{Missing clarification on underspecified turns.}
When the user question is ambiguous, the gold response typically requests clarification.
Our system sometimes commits to an arbitrary interpretation, producing content that is grounded in the passages but non-responsive to the user's actual intent (e.g., \errexref{1}, \errexref{2}, \errexref{3}).
This accounts for the low underspecified-question scores across all models (Table~\ref{tab:category}).

\paragraph{Pragmatically incomplete responses.}
Some responses are faithfully grounded but fail to resolve the user's communicative intent -- for example, omitting an explicit confirmation or contradiction in follow-up turns, or failing to address the user's confusion expressed in prior turns (e.g., \errexref{4}).
These cases score well on faithfulness (RL\_F) but poorly on reference-based metrics (RB\_alg, RB\_llm).

\paragraph{Topic drift under noisy evidence.}
When reference passages contain salient but tangential information, models occasionally drift toward secondary details, yielding grounded but non-responsive answers.

\paragraph{Ungrounded elaboration.}
Occasionally, models extrapolate beyond the provided evidence while maintaining a style similar to the reference answer, resulting in higher reference-based scores but lower faithfulness.

\paragraph{Metric disagreement.}
We observe cases where (i)~faithful responses receive low ROUGE/BERTScore due to missing a single key sentence, and (ii)~the faithfulness judge overestimates scores on off-topic but internally consistent responses (e.g., \errexref{5}).
This motivates interpreting multiple signals jointly rather than relying on any single metric.

\section{Qualitative Error Examples}
\label{sec:qual_errors}

Below we list qualitative examples referenced in the Error Analysis section.

\paragraph{\errex{1} (Underspecified turn: missing clarification)}\label{err:1}
\mbox{}\\
\textbf{User query:} ``When was the season of Grey's Anatomy premiered and when did it conclude?'' \\
\textbf{Evidence:} Contexts describe multiple different seasons (7/9/10/13/14). \\
\textbf{System output:} Lists dates for several seasons instead of asking which season. \\
\textbf{Gold behavior:} Request clarification (which season?).

\paragraph{\errex{2} (Underspecified turn: missing clarification)}\label{err:2}
\mbox{}\\
\textbf{User query:} ``Is there a daily limit to hunt the animal in California?'' \\
\textbf{Evidence:} Bag limits differ by species/category (birds, waterfowl, mammals, wild pig). \\
\textbf{System output:} Enumerates limits for many animals without resolving which animal is meant. \\
\textbf{Gold behavior:} Ask which animal/species the user refers to.

\paragraph{\errex{3} (Underspecified turn: missing clarification)}\label{err:3}
\mbox{}\\
\textbf{User query:} ``Does this outpatient clinic provide mental health services to veterans in Minneapolis?'' \\
\textbf{Evidence:} Context includes multiple clinics/facilities; the referent of ``this clinic'' is ambiguous. \\
\textbf{System output:} Assumes a specific clinic and answers. \\
\textbf{Gold behavior:} Ask which clinic is intended.

\paragraph{\errex{4} (Grounded but pragmatically incomplete)}\label{err:4}
\mbox{}\\
\textbf{User query:} ``What do you mean by `check salary/benefits' page? No fees at my end?'' \\
\textbf{Evidence:} The page states residents receive salary/benefits; some optional tuition/travel may be the resident's responsibility. \\
\textbf{System output:} Mentions salary and possible costs but does not clearly answer ``no fees?'' \\
\textbf{Gold behavior:} Explicitly state that there are no program fees charged to residents (with the caveat about optional expenses).
\paragraph{\errex{5} (Metric disagreement: judge overestimation via unstated generalization)}\label{err:5}
\mbox{}\\
\textbf{User query:} ``I heard the toolchain is not available in South America.'' \\
\textbf{Evidence:} The context states the toolchain is available in \emph{Sao Paulo}. \\
\textbf{System output:} ``The toolchain is available in Sao Paulo.'' \\
\textbf{Gold behavior:} Explicitly contradict the user's claim by connecting the evidence to the claim (e.g., ``That is not true; it is available in Sao Paulo, which is in South America.''). \\
\textbf{Issue:} The model (and the judge) implicitly generalize from ``available in Sao Paulo'' to ``available in South America,'' but the response does not make this link explicit, which can inflate judge-based scores despite a pragmatically incomplete correction.

\section{Detailed Benchmark Analysis}
\label{app:benchmark}

This appendix expands on the benchmark limitations summarized in
Section~\ref{sec:critical}.

\paragraph{Target leakage through empty context.}
All 97 unanswerable questions in the Task~B evaluation set have
\emph{empty} reference passages, creating a trivial shortcut:
detecting empty context and returning ``I don't know'' yields a
perfect conditioned score.
In realistic deployments, unanswerable questions are accompanied by
\emph{irrelevant} retrieved passages, making answerability detection
substantially harder.
We recommend including distractor passages for unanswerable instances.

\paragraph{Annotation coverage limitations.}
Annotators used auto-retrieval and manual search to extract relevant
documents, but these tools may miss relevant content -- especially for
broad questions (e.g., ``How to open a small business?'') where
evidence is scattered or paraphrased.
Consequently, reference answers may be incomplete, potentially
penalizing systems that find additional relevant information.
We recommend employing corpus-level domain experts to validate answer
completeness.

\paragraph{Possible leakage through reference-less metrics.}
The ability to compute the reference-less faithfulness metric (RL\_F)
on the submission set enables selecting the best model per instance
based on this signal -- effectively incorporating evaluation
information into system design.
It remains unclear whether this reflects genuine quality improvement
or partial metric leakage.

\paragraph{Annotation quality.}
We used several LLMs to verify reference answers against provided
contexts and found a notable fraction of partially misleading or
incorrect gold responses (e.g., confusing IBM Cloud bucket retention
times with Watson Assistant retention times).
Current-generation LLMs could serve as validation assistants to flag
such issues, though care must be taken to avoid introducing systematic
biases.

\paragraph{Metric robustness.}
Manual analysis revealed that all metrics occasionally rank models
incorrectly.
Lexical metrics such as ROUGE-L are particularly noisy.
We note that high reported metric--human correlations may be inflated
by the inclusion of both very strong and very weak systems;
correlations among similar-quality systems -- the more practically
relevant scenario -- may be substantially lower.

\paragraph{Guidelines alignment.}
Evaluation would be more robust if generation systems and evaluation
judges operated under identical, explicit guidelines regarding: use of
external world knowledge, handling of corpus--world-knowledge
contradictions, direct vs.\ goal-anticipating responses, and citation
vs.\ synthesis of passages.
The absence of shared guidelines introduces uncontrolled variance into
both generation and evaluation.

\end{document}